\documentclass[conference]{IEEEtran}
\IEEEoverridecommandlockouts
\usepackage{amsmath,amsfonts}
\usepackage{algpseudocode}
\usepackage{algorithm}
\usepackage{array}
\usepackage[caption=false,font=footnotesize,labelfont=sf,textfont=sf]{subfig}
\usepackage{enumitem}
\usepackage{textcomp}
\usepackage{booktabs}
\usepackage{multirow}
\usepackage{stfloats}
\usepackage{url}
\usepackage{verbatim}
\usepackage{graphicx}
\usepackage{tabularx}
\usepackage{cite}
\usepackage{comment}
\usepackage[framemethod=default]{mdframed}
\newmdenv[backgroundcolor=orange, linecolor=orange]{highlightedbox}
\usepackage{xcolor}
\usepackage{soul}
\newcommand{\etal}{\textit{et al.}}

\def\BibTeX{{\rm B\kern-.05em{\sc i\kern-.025em b}\kern-.08em
    T\kern-.1667em\lower.7ex\hbox{E}\kern-.125emX}}
\begin{document}

\title{
FedMFS: Federated Multimodal Fusion Learning with Selective Modality Communication
\thanks{This work is supported by the ONR under Grant N00014-23-C-1016, the NSF under Grant CNS-2212565, and the NSF under Grant CPS-2313109.}
}

\author{
    \IEEEauthorblockN{Liangqi Yuan, Dong-Jun Han, Vishnu Pandi Chellapandi, Stanislaw H. \.{Z}ak, and Christopher G. Brinton\\
    Elmore Family School of Electrical and Computer Engineering, Purdue University, West Lafayette, 47907, USA \\ 
    Email: \{liangqiy, han762, cvp, zak, cgb\}@purdue.edu}}

\maketitle

\begin{abstract}
Multimodal federated learning (FL) aims to enrich model training in FL settings where devices are collecting measurements across multiple modalities (e.g., sensors measuring pressure, motion, and other types of data). However, key challenges to multimodal FL remain unaddressed, particularly in heterogeneous network settings: (i) the set of modalities collected by each device will be diverse, and (ii) communication limitations prevent devices from uploading all their locally trained modality models to the server.
In this paper, we propose \underline{Fed}erated \underline{M}ultimodal \underline{F}usion learning with \underline{S}elective modality communication (FedMFS), a new multimodal fusion FL methodology that can tackle the above mentioned challenges. 
The key idea is the introduction of a modality selection criterion for each device, which weighs (i) the impact of the modality, gauged by Shapley value analysis, against (ii) the modality model size as a gauge for communication overhead.
This enables FedMFS to flexibly balance performance against communication costs, depending on resource constraints and application requirements. 
Experiments on the real-world ActionSense dataset demonstrate the ability of FedMFS to achieve comparable accuracy to several baselines while reducing the communication overhead by over 4x.
\end{abstract}

\section{Introduction}
\label{Sec. Introduction}

Federated learning (FL) is a distributed machine learning (ML) approach in which users collaboratively train an ML model through sharing model parameters rather than raw measurements~\cite{mcmahan2017communication,yuan2023decentralized}. The conventional FL paradigm involves clients training ML models on local data, then uploading them to a central server for aggregation, and synchronizing the devices to begin the next training round~\cite{fang2022communication,chellapandi2023convergence}. As edge devices in the Internet of Things (IoT), such as smartphones, robots, unmanned aerial vehicles (UAVs), are often equipped with multimodal sensors, there has been an increasing interest in multimodal fusion federated learning (MFFL) frameworks. For example, consider a set of connected and automated vehicles (CAVs) with sensors such as cameras, LiDAR, and Radar~\cite{chellapandi2023survey,chellapandi2023federated}. These multimodal sensors enable control decisions in various driving scenarios, including different weather conditions and fields of view. CAVs are expected to rely on MMFL to collaboratively learn ML models across vehicles~\cite{yuan2023peer}.

\noindent \textbf{Relevant Literature.} Recently, a variety of algorithms have been proposed to improve the performance of MFFL based on different fusion techniques. Qi \etal~\cite{qi2023fl} proposed a data-level fusion FL system tailored for the combination of wearable sensor signals and images for user fall detection. Xiong \etal~\cite{xiong2022unified} implemented a feature-level fusion approach with attention modules. Feng \etal~\cite{feng2023fedmultimodal} presented two decision-layer fusion strategies using concatenation and attention modules, respectively, to address three types of client heterogeneities: modality absence, label absence, and label errors. Outside of fusion strategies, Salehi \etal~\cite{salehi2022flash} incorporated the MFFL framework into CAVs and conducted real-world experiments. In their FLASH framework, clients randomly select and upload one of the three modality models or an ensemble model for aggregation. Chen \etal~\cite{chen2022fedmsplit} integrated MFFL into decentralized FL, aiming to facilitate collaborative training within client networks without server support. 

\begin{figure}[t]
\centering
\includegraphics[width=\linewidth]{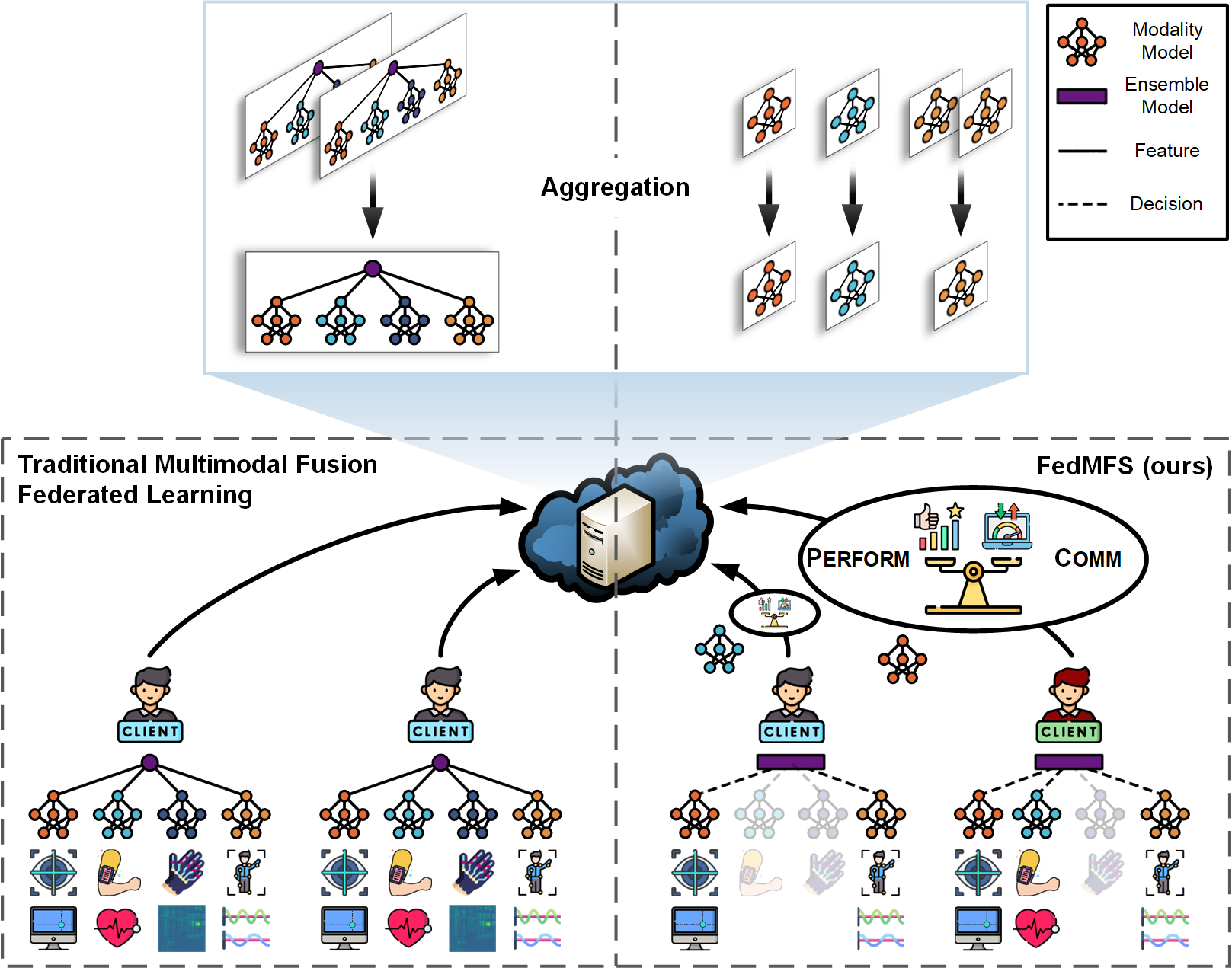}
\caption{Schematic representation of traditional multimodal fusion federated learning vs. the proposed FedMFS.}
\label{Fig. FedMFS}
\vspace{-10pt}
\end{figure}

\noindent \textbf{Problem Statement.} In the context of IoT, where devices measure a diverse set of modalities, most existing MFFL frameworks call for massive parallel processing of extensive sensor data. They utilize multimodal fusion to boost model performance, especially in scenarios where heterogeneous clients lack certain modalities. However, IoT devices often possess communication constraints due to bandwidth limitations and diverse capabilities, which has been a key bottleneck in conventional (single modality) FL~\cite{imteaj2021survey,fang2023submodel}. Thus, there is still a need to devise strategies to reduce communication overhead and improve learning efficiency within the MFFL paradigm.
Motivated by this, we aim to answer the following key question: 
\vspace{5pt}
\begin{quote}
\textit{\textbf{In resource-constrained and heterogeneous MFFL settings, how should each client evaluate and select the best set of modalities to trade-off between performance and communication?}}
\end{quote}

\noindent \textbf{Overview.} An overview of our solution is given in Fig.~\ref{Fig. FedMFS}. We propose a decision-level fusion approach, where predictions from \textit{modality models} are used as inputs to the \textit{ensemble model}. This allows the independent deployment of the modality models in various application scenarios, accommodating situations where each client may possess a different set of modalities. Furthermore, we consider classical ML models for our ensemble instead of the complex neural networks typically used in traditional MFFL, resulting in reduced computational overhead and improved interpretability. Lastly, we introduce \textit{selective modality communication} to account for the fact that clients do not always have the ability to upload all modality models. To measure the impact of each modality, we propose leveraging Shapley values~\cite{sundararajan2020many,lundberg2017unified,lundberg2020local2global} measured on the ensemble model to quantify their respective performances. Recent studies using Shapley values have demonstrated that, in data/sensor fusion, different input features have varying impacts on the model output~\cite{yuan2022interpretable,yuan2023three,yuan2023passive,yang2023extraction}. 

Although many communication-efficient FL frameworks have been developed~\cite{shah2021model,yuan2023federated,wu2022communication}, our focus is on specific MFFL scenarios where further reduction in communication overhead is necessary. Our approach can be applied on top of these other frameworks.

\noindent \textbf{Contributions.} We make the following contributions: 
\begin{itemize} 
    \item We propose \underline{Fed}erated \underline{M}ultimodal \underline{F}usion learning with \underline{S}elective modality communication (FedMFS), an MMFL methodology tailored for clients with heterogeneously absent modalities. A Shapley component incorporated within FedMFS enables quantifying and interpreting the impact of individual modalities. 
    \item We provide a series of flexible configurations to balance the performance of the ML model with the communication overhead. Specifically, through three customizable parameters, FedMFS aims to minimize communication costs while ensuring optimal learning efficiency, depending on the particular application and available communication resources. The FedMFS framework also facilitates a modular architecture for the modality models, allowing them to operate independently.
    \item We evaluate the performance and communication overhead of our proposed FedMFS against four baseline methods on the heterogeneous multimodal dataset ActionSense. Experimental results demonstrate the superior performance of FedMFS, which, while achieving comparable accuracy, incurs less than 25\% of the communication overhead compared to the baselines. 
\end{itemize}

\section{Formulation and Methodology}
\label{Sec. Methodologies}

\subsection{Federated Learning and Fusion Setup} 
We assume that there are $K$ clients participating in the FL framework. Each client, denoted $k$, possesses a dataset $\mathbb{D}^k$ and a label set $Y^k$, comprising multimodal data represented as
\begin{equation}
    \mathbb{D}^k = \{\mathcal{D}^k_1, \mathcal{D}^k_2, \dots, \mathcal{D}^k_{M_k}\},
\end{equation}
where $\mathcal{D}^k_m$ denotes datasets corresponding to modality $m$, such as images, LiDAR, RF, etc., with $m = 1, 2, \dots, M_k$ indicating specific data modality. We note that the heterogeneous client $k$ can accommodate a different number of data modalities, represented by $M_k$. Each client has a \textit{modality model}, $\theta^k_m$, for every modality dataset $\mathcal{D}^k_m$, which is designed to capture the relationship between the input data and its corresponding labels. Therefore, each client possesses a set of models, represented as
\begin{equation}
    \Theta^k = \{\theta^k_1, \theta^k_2, \ldots, \theta^k_{M_k}\}.
\end{equation}
Each model, when trained on the dataset, yields a predicted label, denoted as
\begin{align}
    \hat{y}^k_m &= \theta^k_m(\mathcal{D}^k_m), \\
    \widehat{\mathbb{Y}}^k &= \{\hat{y}^k_1, \hat{y}^k_2, \dots, \hat{y}^k_{M_k}\},
\end{align}
where $\hat{y}^k_m$ represents the predicted label generated from the model $\theta^k_m$ for the dataset $\mathcal{D}^k_m$, and $\widehat{\mathbb{Y}}^k$ is the collection of all the predicted labels for client $k$. Our goal is to fuse the outputs of all modality models at the decision layer through a post-processing \textit{ensemble model} $\omega^k$, represented as
\begin{align}
    \widehat{Y}^k &= \omega^k \left(\theta^k_1(\mathcal{D}^k_1), \theta^k_2(\mathcal{D}^k_2), \ldots, \theta^k_{M_k}(\mathcal{D}^k_{M_k})\right) \nonumber \\
                  &= \omega^k \left(\hat{y}^k_1, \hat{y}^k_2, \ldots, \hat{y}^k_{M_k}\right) \nonumber \\
                  &= \omega^k \left(\widehat{\mathbb{Y}}^k\right),
\end{align}
where $\widehat{Y}^k$ denotes the predicted label set corresponding to the dataset $\mathbb{D}^k$. 

\subsection{Proposed FedMFS Algorithm}
The proposed FedMFS is described in Algorithm~\ref{Alg. FedMFS}. The primary objective of FedMFS is the collaborative learning of the modality model $\theta_m$ for each modality $m$. For each client, individual and system heterogeneities (e.g., modality missing, noise, device errors, device malfunctions, etc.) are addressed through a personalized ensemble model $\omega^k$. Furthermore, an intrinsic objective of FedMFS is to compensate for the communication constraints inherent in IoT edge devices by minimizing communication overhead, thus ensuring efficient learning efficacy for clients within the FL framework.

\renewcommand{\Procedure}[2]{\item[\hspace{0em}\textbf{#1}] #2}
\renewcommand{\EndProcedure}{}
\makeatletter
\newcommand{\setalglineno}[1]{%
  \setcounter{ALG@line}{\numexpr#1-1}}
\makeatother

\begin{algorithm}[h]
\small
\caption{FedMFS: Federated Multimodal Fusion Learning with Selective Modality Communication}
\label{Alg. FedMFS}
\textbf{Input:} Communication rounds ($T$), clients' dataset ($\mathbb{D}^k$), clients' label set ($Y^k$), local training epoch ($E$), initial models ($\theta^k_{m,0}$ \& $\omega^k_0$), loss function ($\mathcal{L}$), learning rate ($\eta$), modality model upload count ($\gamma$), performance and communication weights ($\alpha_s$ \& $\alpha_c$)

\textbf{Output:} Generalized global modality models ($\theta_m$) and personalized local ensemble models for each client ($\omega^k$)

\begin{algorithmic}[1]
\Procedure{\# Global Iteration}{}
\For{$t=0$ {\bfseries to} $T-1$}
    \begin{algorithmic}[1]
    \Procedure{}{}
    
    \begin{algorithmic}[1]
    \Procedure{\# Local Learning}
    \For{each client $k$ {\bfseries in parallel}}
        \For{each data modality $m$}
            \State Backpropagate the loss function and update the modality models $\theta^{k,t}_m \gets \arg\min_{\theta^{k,t}_m} \mathcal{L}(Y^{k,t}, \hat{y}^{k,t}_m)$.
        \EndFor
        \State (Stage \#1) Update the ensemble model $\omega^{k,t} \gets \arg\min_{\omega^{k,t}} \mathcal{L}(Y^{k,t}, \widehat{Y}^{k,t})$.
    \EndFor
    \EndProcedure
    \end{algorithmic}

    \begin{algorithmic}[1]
    \Procedure{\# Performance-Communication Trade-Off}{}
    \State Compute the impact of each modality $\Phi^{k,t}$ on the prediction using the Shapley values. \Comment{(\ref{Eq. Shapley}), (\ref{Eq. Shapley Value Set})}
    \State Compute the modality model size $\Bar{\Theta}^k$. \Comment{(\ref{Eq. Modality Model Size})}
    \State Normalize and select top-$\gamma$ priority modality models for uploading. \Comment{(\ref{Eq. Normalization}), (\ref{Eq. Priority}), (\ref{Eq. Top Priority})}
    \EndProcedure
    \end{algorithmic}	

    \begin{algorithmic}[1]
    \Procedure{\# Server Aggregation}{}
    \For{each data modality $m$}
        \State Server calculates the weight $\beta^{k,t}_m$ for each client $k$ and aggregates modality model $\theta^{t}_m$. \Comment{(\ref{Eq. Aggregation}), (\ref{Eq. Aggregation Weight})} 
    \EndFor
    \EndProcedure
    \end{algorithmic}
    
    \begin{algorithmic}[1]
    \Procedure{\# Local Deploying}{}
    \For{each client $k$ {\bfseries in parallel}}
        \For{each data modality $m$}
            \State Deploy the downloaded global modality model $\theta^{t}_m$.
            \State Re-calculate $\widehat{Y}^{k,t}$ based on the global modality model $\theta^{t}_m$.
        \EndFor
        \State (Stage \#2) Update the ensemble model $\omega^{k,t} \gets \arg\min_{\omega^{k,t}} \mathcal{L}(Y^{k,t}, \widehat{Y}^{k,t})$.
    \EndFor
    \EndProcedure
    \end{algorithmic}
    
    \EndProcedure
    \end{algorithmic}
    
\setalglineno{2}
\State \hspace{-0.5em} Update the modality models for each modality $\theta_m \gets \theta^{T}_{m}$.
\State \hspace{-0.5em} Update the ensemble models for each client $\omega^{k} \gets \omega^{k,T}$.
\State \hspace{-2em} \textbf{end for}

\end{algorithmic}	
\end{algorithm}

\subsection{Client Learning} 

For each data modality for every client, the local objective is to minimize the difference between the predicted and the true labels. This objective can be achieved using various optimizers, such as stochastic gradient descent (SGD). Formally, we describe the optimization problem as $ \min_{\theta^k_m} \mathcal{L}(Y^k, \hat{y}^k_m)$,   
where $\mathcal{L}$ is a loss function that measures the discrepancy between the true label set and the predicted label set. For the ensemble model $\omega^k$, the learning objective is to minimize the discrepancy between the true label set and the predicted label sets across all modalities, as $\min_{\omega^k} \mathcal{L}(Y^k, \widehat{Y}^k)$.

\addtolength{\topmargin}{0.05in}

\subsection{Performance-Communication Trade-Off} 
Due to the resource constraints on edge devices serving as clients, they may not possess adequate storage capacity to house extensive multimodal data, computational capability to learn on multimodal datasets, or communication bandwidth to upload several models to the server. Thus, we introduce two metrics to assist clients to determine whether to upload their models to the server:
\begin{itemize}
    \item \textbf{Shapley value} ($\varphi$) represents the impact of a modality model on the final prediction, where a higher value of $\varphi$ is preferred ($\uparrow$).
    \item \textbf{Modality model size} ($|\theta|$) pertains to the communication overhead, where a lower value of $|\theta|$ is preferred ($\downarrow$).
\end{itemize}

\noindent \textbf{Shapley Value (Impact).} During each communication round, clients evaluate the impact of the models $\Theta^k$ on the outcomes utilizing interpretability techniques and choose to upload only a singular selected model $\theta^k$. We consider using the Shapley value as an assessment to evaluate the relationship between input $\widehat{\mathbb{Y}}^k$ and output $\widehat{Y}^k$ of the ensemble model $\omega^k$:
\begin{equation}
    \begin{small}
    \begin{aligned}
    \varphi^k_m = \sum_{\mathcal{Y} \subseteq \widehat{\mathbb{Y}}^k \setminus \{m\}} \frac{|\mathcal{Y}|!(|\widehat{\mathbb{Y}}^k|-|\mathcal{Y}|-1)!}{|\widehat{\mathbb{Y}}^k|!} \left( \omega^k(\mathcal{Y} \cup \{m\}) - \omega^k(\mathcal{Y}) \right),
    \end{aligned}
    \end{small}
\label{Eq. Shapley}
\end{equation}
where $\varphi^k_m$ is the Shapley value of input modality $m$, $\mathcal{Y}$ is a subset of $\widehat{\mathbb{Y}}^k$ excluding modality $m$, and $\omega^k(\mathcal{Y})$ is the predicted value using only modalities in set $\mathcal{Y}$. For all modalities, we assess the magnitude of each Shapley value by taking its absolute value and construct the following set:
\begin{equation}
    \Phi^k = \left\{|\varphi^k_1|, |\varphi^k_2|, \dots, |\varphi^k_{M_k}|\right\}.
\label{Eq. Shapley Value Set}
\end{equation}

\noindent \textbf{Modality Model Size (Communication Overhead).} Given modality models with parameters $\Theta^k = \{\theta^k_1, \theta^k_2, \ldots, \theta^k_{M_k}\}$, the communication overhead for each modality model is directly proportional to the model size given by
\begin{equation}
\Bar{\Theta}^k = \{|\theta^k_1|, |\theta^k_2|, \ldots, |\theta^k_{M_k}|\}. 
\label{Eq. Modality Model Size}
\end{equation}

\noindent \textbf{Priority (Composite Score).} Considering the impact of the modality model, as quantified by the Shapley value and the communication overhead as characterized by the modality model size, we propose priority $P$ as a composite score. To derive the priority, we proceed with individual normalization for each criterion:
\begin{equation}
    \begin{small}
    \left\{\!
    \begin{aligned}
    \tilde{\varphi}^k_m &= \frac{\varphi^k_m - \min(\Phi^k)}{\max(\Phi^k) - \min(\Phi^k)}, \\
    |\tilde{\theta}^k_m| &= \frac{\theta^k_m - \min(\Bar{\Theta}^k)}{\max(\Bar{\Theta}^k) - \min(\Bar{\Theta}^k)}, 
    \end{aligned}
    \right. \text{for } m = 1, 2, \dots, M_k,
    \end{small}
\label{Eq. Normalization}
\end{equation}
where $\tilde{\varphi}^k_m$ represents the normalized Shapley Value and $|\tilde{\theta}^k_m|$ denotes the normalized communication overhead. With a focus on identifying modality models for server communication, we devise the priority $P^k_m$ for each modality and the corresponding set $\mathcal{P}^k$ to determine whether modality models should be sent to the server. They are formulated as
\begin{equation}
\begin{aligned}
    P^k_m &= \alpha_s \times \tilde{\varphi}^k_m + \alpha_c \times (1 - |\tilde{\theta}^k_m|), \\
    \mathcal{P}^k &= \left\{P^k_1, P^k_2, \dots, P^k_{M_k}\right\},
\end{aligned}
\label{Eq. Priority}
\end{equation}
where $\alpha_s$ and $\alpha_c$ are the predetermined metric weights, satisfying $\alpha_s + \alpha_c = 1$. Naturally, a modality with the maximal priority is considered optimal.

To streamline our decision-making, we focus on modalities with scores among the top-$\gamma$ priority:
\begin{align}
    \mathcal{P}^k_\gamma &= \mathrm{top}\max_\gamma (\mathcal{P}^k) \nonumber \\
    &= \{ x \,:\, x \in \mathcal{P}^k \, \text{and} \, |\mathcal{P}^k \cap \{y \,|\, y \geq x\}| \leq \gamma \}.
\label{Eq. Top Priority}
\end{align}
Hence, the set of modalities that client $k$ communicates to the server becomes:
\begin{equation}
\begin{small}
    \Theta^k_\gamma = \left\{\theta^k_m : \theta^k_m \in \Theta^k \, \text{and} \, P^k_m \in \mathcal{P}^k_\gamma, \text{ for } m = 1, 2, \dots, M_k\right\},
\end{small}
\label{Eq. Positive Shapley}
\end{equation}
where the set $\Theta^k_\gamma$ represents the modality models corresponding to the top-$\gamma$ priority from $\mathbb{S}^k$. Each client will upload a data packet with various details to the server for aggregation, including model parameters $\theta^k$, modality information $m$, the number of samples $|\mathcal{D}^k_m|$, among others. Likewise, upon downloading from the server, this information will also be retrieved. Note that only the model $\theta^k$ will be uploaded/downloaded to/from the server. The ensemble model $\omega^k$ varies across clients, determined by the unique deployment scenarios of each client, such as geographical location, operational duration, external interference, etc.

\subsection{Model Aggregation}
Upon receiving data packets from the clients, the server performs a weighted aggregation of the models based on the number of samples for each data modality. For a given data modality $m$, the server aggregates the model parameters from client $k$ with modality $m$. The update is given by
\begin{equation}
    \theta_m \leftarrow \sum_{\theta^k_m \in \Theta^k_p} \beta^k_m \theta^k_m,
\label{Eq. Aggregation}
\end{equation}
where $\beta^k_m$ represents the aggregation weight coefficient. Following the methodology adopted in FedAvg~\cite{mcmahan2017communication}, these weights are determined based on the number of samples, and can be expressed as
\begin{equation}
    \beta^k_m = \frac{|\mathcal{D}^k_m|}{\sum_{k=1}^{K_m}|\mathcal{D}^k_m|},
\label{Eq. Aggregation Weight}
\end{equation}
where $K_m$ denotes the number of client models received by the server for modality $m$. 

\subsection{Deployment}
In the FedMFS framework, only the modality models $\theta_m$ are aggregated, while the ensemble models $\omega_k$ remain separate. The purpose is to give the modality models a global perspective, ensuring a wider generalization. These modality models adhere to the classical FL iterative process and are deployed as global modality models. On the contrary, the personalized ensemble model undergoes a two-stage update. In Stage \#1, the ensemble model serves as an intermediate state, primarily facilitating the calculation of the Shapley values. It is not deployed in any application (i.e., it is not tested on any test set). In Stage \#2, after receiving the global modality models from the server, the clients subsequently update their ensemble model using global modality models in conjunction with local data to achieve the final state.

\section{Experiment and Results}

\subsection{Experiment Setup}

\noindent\textbf{Dataset.}
We use the multimodal dataset, ActionSense~\cite{delpreto2022actionsense}, to validate the proposed FedMFS. ActionSense is a comprehensive multimodal dataset that captures human daily activities, integrating various wearable and environmental sensors. It captures data of human interactions with objects and the environment in a kitchen setting as tasks are performed. This experiment illustrates the application of FL leveraging wearable sensors to implement FL and protect user privacy. We utilize six modalities of sensors from ActionSense, with their descriptions shown in Table \ref{Table Dataset}. To ensure fairness with comparison, we adopted the sample code provided by ActionSense for preprocessing sensor data, including filtering, resampling (since the sensors have distinct sampling rates), normalization, and so forth. 

\begin{table}[h] 
\caption{Description of ActionSense Dataset}
\label{Table Dataset}
\centering
\begin{tabular}{@{}lccccc@{}}
\toprule
\multirow{2}{*}{\textbf{Sensor}} & \multirow{2}{*}{\textbf{Type}} & \multirow{2}{*}{\textbf{Position}} & \multirow{2}{*}{\textbf{Feature}} & \textbf{Heterogeneity} \\
 & & & & \textbf{(Missing Data)} \\

\midrule
Eye Tracking & Position & Head & $2$ & \\
Myo & EMG & Left Arm & $8$ & \\
Myo & EMG & Right Arm & $8$ & \\
Tactile Glove & Pressure & Left Hand & $32 \times 32$ & S06 -- S09 \textsuperscript{1} \\
Tactile Glove & Pressure & Right Hand & $32 \times 32$ & S06 -- S09 \textsuperscript{1} \\
Xsens & Rotation & Body & $22 \times 3$ & \\
\bottomrule
\end{tabular}

\vspace{2pt}
\raggedright
\noindent{\textsuperscript{1} S06 -- S09 refers to subjects 06 through 09.}

\end{table}

\noindent\textbf{Base Models.}
To ensure a fair comparison, for all the aforementioned data modalities, we initially reshape them into two dimensions, i.e., time $\times$ features. We employ a consistent Long Short-Term Memory (LSTM) network structure for all modalities, comprising a single LSTM layer with $64$ hidden units, followed by a fully connected layer, and using a LogSoftmax for output. We adopt negative log likelihood loss (NLLLoss), with SGD as optimizer, a learning rate $\eta = 0.1$, a batch size of $32$, a local training epoch $E = 5$, and a total of $T = 100$ communication rounds. Four base models for the six modalities; Eye, Myo, Tactile, and Xsens have sizes of 0.07 MB, 0.08 MB, 1.07 MB, and 0.13 MB, respectively.

\noindent\textbf{Baselines.}
At the system level, we consider three traditional multimodal sensor fusion frameworks for FL, encompassing data-level~\cite{qi2023fl}, feature-level~\cite{xiong2022unified}, and decision-level~\cite{feng2023fedmultimodal} fusion as our baselines. Additionally, FLASH~\cite{salehi2022flash} with uniform model selection probabilities serves as another baseline. To ensure a fair systematic comparison within the FL context, we do not incorporate various specialized techniques present in these baselines, such as co-attention mechanisms. All these baselines employ a uniform network architecture, specifically, an LSTM layer followed by a fully connected layer, utilizing a concatenate strategy for fusion.

\noindent\textbf{Proposed FedMFS.} For the proposed FedMFS framework, in addition to the fundamental configurations mentioned above, these modality models do not output logarithmic probabilities but rather provide definitive predicted categories ($\widehat{\mathbb{Y}}$) for the ensemble model ($\omega$). The ensemble model can adopt various choices depending on the specific use case and the resources available on the client, such as voting methods, linear models, \textit{k}-nearest neighbors (\textit{k}-NN), etc. Here, we use the Random Forest (RF) as our ensemble model because of its robust interpretability. We perform a subsampling on the dataset, selecting $50$ samples to compute the Shapley values to reduce computational complexity. Note that the ensemble model, Shapley values, as well as the modality model sizes are kept private by the client and used for modality model selection. They are neither uploaded to the server nor does the server possess knowledge of the client's computing methodology.

\begin{table}[t]
\caption{Comparison of Accuracy and Communication Overhead at Cumulative Consumption of 50 MB}
\label{Table Experimental results}
\centering 
\scriptsize
\begin{tabular}{@{}lcccccc@{}}
\toprule
\multirow{2}{*}{Method} & \multirow{2}{*}{$\gamma$} & \multirow{2}{*}{$\alpha_s$} & \multirow{2}{*}{$\alpha_c$} &  Acc. \textsuperscript{1} & Comm. \textsuperscript{2} & Comm. \\
& & & & (\%) ($\uparrow$) & (MB) ($\downarrow$) & Round \\
\midrule
Data-level~\cite{qi2023fl} & & & & 47.89 & 19.36 & 2 \\
Feature-level~\cite{xiong2022unified} & & & & 50.45 & 13.97 & 3 \\
Decision-level~\cite{feng2023fedmultimodal} & & & & 36.44 & 14.01 & 3 \\
FLASH~\cite{salehi2022flash} & & & & 54.81 & 2.08 & 24 \\
\midrule

\multirow{30}{50pt}{\textbf{FedMFS proposed by us}} 
& \multirow{5}{*}{\textbf{1}} & 1 & 0 & 82.90 & 3.39 & 14 \\
& & 0.8 & 0.2 & 81.98 & 2.68 & 19 \\
& & 0.5 & 0.5 & 95.95 & 0.85 & 60 \\
& & \textbf{0.2} & \textbf{0.8} & \textbf{97.34} & \textbf{0.72} & \textbf{69} \\
& & 0 & 1 & 70.43 & 0.64 & 77 \\
\cmidrule{2-7}
& \multirow{5}{*}{2} & 1 & 0 & 79.52 & 5.55 & 8 \\
& & 0.8 & 0.2 & 86.94 & 4.63 & 11 \\
& & 0.5 & 0.5 & 92.36 & 1.67 & 29 \\
& & 0.2 & 0.8 & 90.31 & 1.55 & 32 \\
& & 0 & 1 & 81.62 & 1.34 & 37 \\
\cmidrule{2-7}
& \multirow{5}{*}{3} & 1 & 0 & 82.61 & 7.22 & 6 \\
& & 0.8 & 0.2 & 86.60 & 6.50 & 8 \\
& & 0.5 & 0.5 & 87.89 & 2.55 & 19 \\
& & 0.2 & 0.8 & 85.48 & 2.23 & 22 \\
& & 0 & 1 & 81.94 & 2.03 & 24 \\
\cmidrule{2-7}
& \multirow{5}{*}{4} & 1 & 0 & 80.68 & 8.43 & 5 \\
& & 0.8 & 0.2 & 85.06 & 8.11 & 5 \\
& & 0.5 & 0.5 & 83.35 & 5.52 & 8 \\
& & 0.2 & 0.8 & 83.15 & 3.24 & 15 \\
& & 0 & 1 & 84.22 & 3.24 & 15 \\
\cmidrule{2-7}
& \multirow{5}{*}{5} & 1 & 0 & 82.05 & 8.93 & 5 \\
& & 0.8 & 0.2 & 82.65 & 10.98 & 4 \\
& & 0.5 & 0.5 & 83.21 & 8.57 & 5 \\
& & 0.2 & 0.8 & 82.74 & 8.58 & 5 \\
& & 0 & 1 & 81.58 & 8.58 & 5 \\
\cmidrule{2-7}
& \multirow{5}{*}{6} & 1 & 0 & 81.86 & 9.16 & 5 \\
& & 0.8 & 0.2 & 74.88 & 13.93 & 3 \\
& & 0.5 & 0.5 & 73.61 & 13.93 & 3 \\
& & 0.2 & 0.8 & 71.84 & 13.93 & 3 \\
& & 0 & 1 & 72.69 & 13.93 & 3 \\
\bottomrule
\end{tabular}

\vspace{2pt}
\raggedright
\noindent{\textsuperscript{1} Average accuracy between clients, $\uparrow$ refers to the higher (preferred).}

\noindent{\textsuperscript{2} Communication overhead per iteration, $\downarrow$ refers to the lower (preferred).}
\end{table}

\subsection{Results}

\begin{figure}[t]
\centering
\includegraphics[width=1\linewidth]{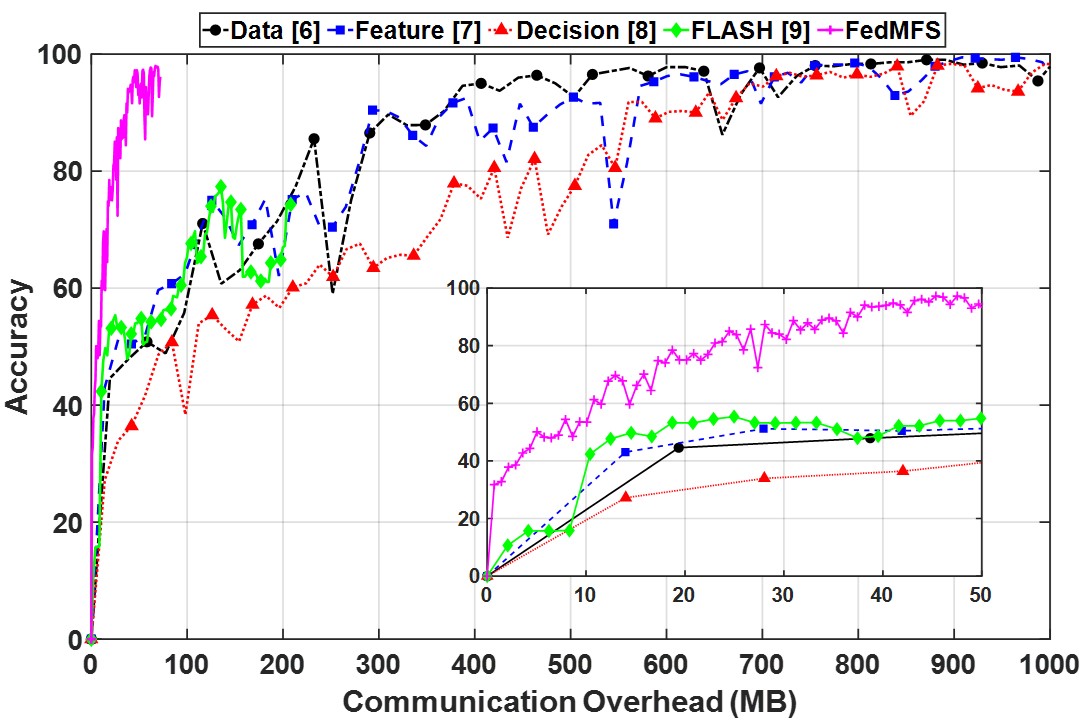}
\caption{Comparison of accuracy between FedMFS with the configuration $\gamma=1, \alpha_s=0.2, \alpha_c=0.8$ and four baselines on a communication overhead scale. Only up to 1000 MB of communication overhead is depicted, while the data-level fusion approach requires close to 2000 MB to complete all iterations.}
\label{Fig. Acc_vs_Comm} 
\end{figure}

The results of the proposed FedMFS, in comparison with four baselines on the ActionSense dataset, are presented in Table \ref{Table Experimental results} and Fig.~\ref{Fig. Acc_vs_Comm}.

\noindent\textbf{Trade-Off Analysis.} Considering the communication constraints, our results underscore the need to find a compromise between $\alpha_s$ and $\alpha_c$ to optimize performance when $\gamma$ is constant. Increasing $\gamma$ does not always lead to better results, as it can exacerbate communication overhead. In some instances, communicating only a select few informative modalities yields enhanced performance. In this context, the proposed FedMFS framework facilitates a flexible determination of the number of modality models to upload, balancing model performance, communication overhead, and learning efficiency.

Specifically for ActionSense, the configuration of $\gamma=1, \alpha_s=0.2, \alpha_c=0.8$ yields the best accuracy with the least communication overhead. This configuration leads most clients to predominantly upload modalities such as Eye Tracking, Myo--Right, and Xsens, which are characterized by fewer input features and thus they are more compact. From extensive experimentation, we observed that in the initial stages of FL, the Eye Tracking modality consistently had a higher upload frequency due to its minimal modality model size, granting it an advantageous position in the trade-off. However, as communication rounds progress, the selection frequency of Eye Tracking decreases, while that of Myo--Right or Xsens increases. This trend can be attributed to the fact that, although the Eye Tracking modality results in the least communication overhead, its limited feature set captures a lower degree of information, giving a slight lag in recognition accuracy.

\begin{figure}[t]
\centering
\includegraphics[width=1\linewidth]{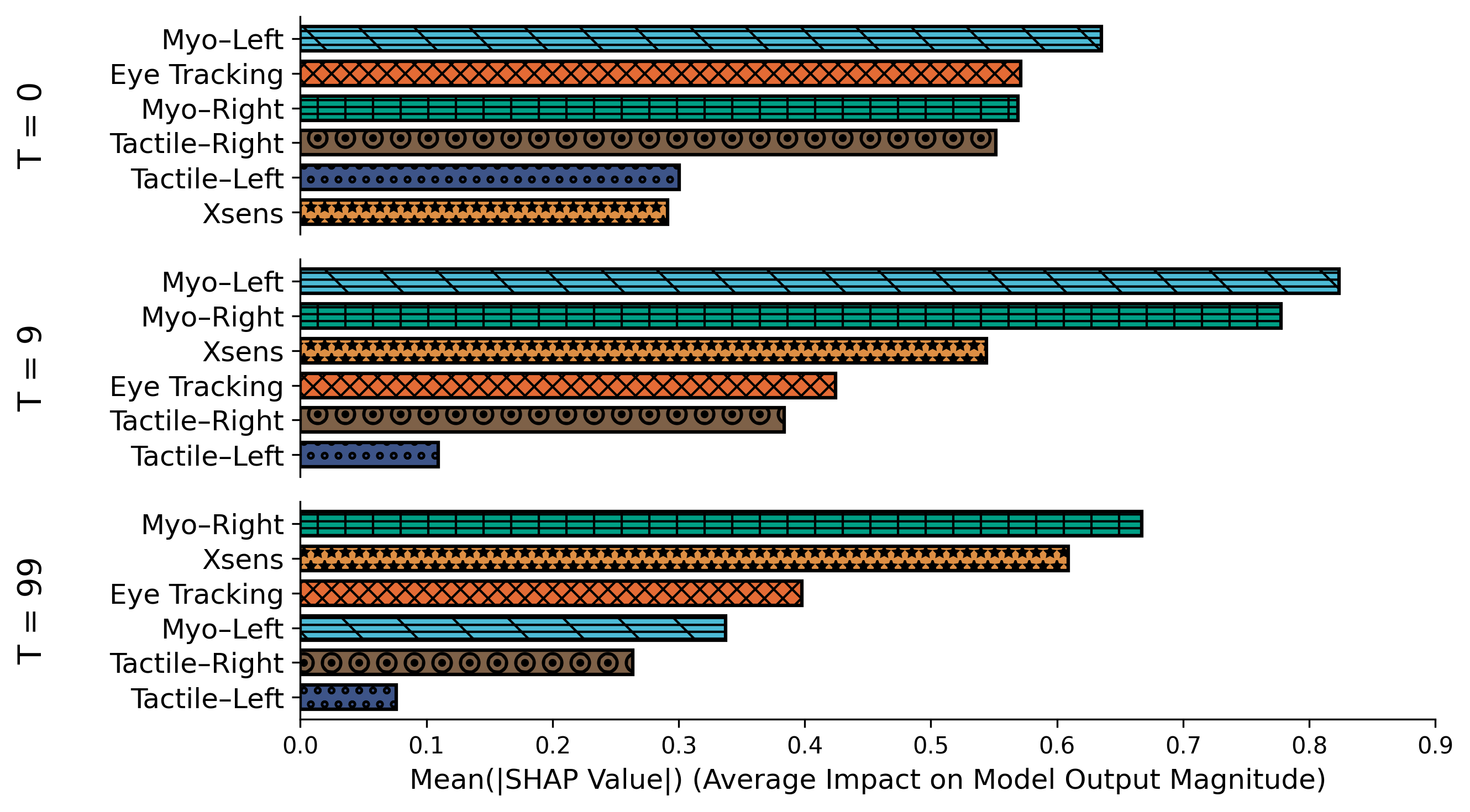}
\caption{The impact of modality models on the ensemble model's final prediction throughout the FedMFS iteration, exemplified with the configuration $\gamma=1, \alpha_s=0.2, \alpha_c=0.8$.}
\label{Fig. Interpretability} 
\end{figure}

\noindent\textbf{Comparison with Baselines.} The proposed FedMFS not only exceeds the four baselines in terms of accuracy but also manages to reduce communication overhead by nearly an order of magnitude, thanks to its selective upload strategy. Insights obtained from FedMFS's selection approach are:
\begin{enumerate}[label=(\roman*)]
    \item Different data modalities contribute distinctively to recognition accuracy.
    \item Aggregation of all modal models is not always necessary.
    \item Utilizing all modalities for fusion is not imperative.
\end{enumerate}
Although the FLASH-adopted random upload strategy effectively reduces communication costs to approximately $\frac{1}{M_k+1}$, it lacks a performance and communication overhead-based selection mechanism, leading to suboptimal results. It is worth noting that the sizes of the modality models vary, which accounts for the reduced communication overhead of FedMFS compared to the FLASH framework. Fig.~\ref{Fig. Acc_vs_Comm} illustrates that FedMFS can achieve much faster convergence with minimal communication overhead, superior performance, and high learning efficiency.

\noindent\textbf{Interpretability of FedMFS.} In addition to helping clients select the modality models, the Shapley value also offers an interpretative approach to quantify the modality models. During the FL process, we can see the clients' favor and the efficacy of each modality model within the FedMFS framework. Fig.~\ref{Fig. Interpretability} illustrates this dynamics, showing the impact of each data modality on the final prediction of the ensemble model across different communication rounds $T$. In the initial stages of FL, most modalities exhibit similar impacts. As communication rounds progress, modalities with larger feature sets and complex models, due to their higher communication overhead, take on a subordinate role in their selection within FedMFS. As FL advances, more straightforward modalities that still convey ample information, such as Myo--Right, emerge as primary contributors.

\section{Conclusion}

In this paper, we presented the FedMFS framework, leveraging Shapley values and modality model sizes to quantify the performance and communication overhead of each modality. In addition to achieving considerable recognition accuracy and reducing communication costs by almost one twentieth, the proposed FedMFS is suitable for heterogeneous clients, features detachable modular modality models, and offers interpretability for data modalities. Our future work will focus on enhancing the adaptability of FedMFS with customizable configurations to fully exploit scenarios where clients might possess dynamic communication capabilities, such as higher bandwidth. Furthermore, Shapley values can also aid in refining the training process of modality models, for example, by potentially discarding underperforming modalities like Myo--Left, thus optimizing computational efficiency.

\bibliographystyle{IEEEtran}
\small\bibliography{reference}

\begin{thebibliography}{10}
\providecommand{\url}[1]{#1}
\csname url@samestyle\endcsname
\providecommand{\newblock}{\relax}
\providecommand{\bibinfo}[2]{#2}
\providecommand{\BIBentrySTDinterwordspacing}{\spaceskip=0pt\relax}
\providecommand{\BIBentryALTinterwordstretchfactor}{4}
\providecommand{\BIBentryALTinterwordspacing}{\spaceskip=\fontdimen2\font plus
\BIBentryALTinterwordstretchfactor\fontdimen3\font minus
  \fontdimen4\font\relax}
\providecommand{\BIBforeignlanguage}[2]{{%
\expandafter\ifx\csname l@#1\endcsname\relax
\typeout{** WARNING: IEEEtran.bst: No hyphenation pattern has been}%
\typeout{** loaded for the language `#1'. Using the pattern for}%
\typeout{** the default language instead.}%
\else
\language=\csname l@#1\endcsname
\fi
#2}}
\providecommand{\BIBdecl}{\relax}
\BIBdecl

\bibitem{mcmahan2017communication}
B.~McMahan, E.~Moore, D.~Ramage, S.~Hampson, and B.~A. y~Arcas,
  ``Communication-efficient learning of deep networks from decentralized
  data,'' in \emph{Artificial intelligence and statistics}.\hskip 1em plus
  0.5em minus 0.4em\relax PMLR, 2017, pp. 1273--1282.

\bibitem{yuan2023decentralized}
L.~Yuan, L.~Sun, P.~S. Yu, and Z.~Wang, ``Decentralized federated learning: A
  survey and perspective,'' \emph{arXiv preprint arXiv:2306.01603}, 2023.

\bibitem{fang2022communication}
W.~Fang, Z.~Yu, Y.~Jiang, Y.~Shi, C.~N. Jones, and Y.~Zhou,
  ``Communication-efficient stochastic zeroth-order optimization for federated
  learning,'' \emph{IEEE Transactions on Signal Processing}, vol.~70, pp.
  5058--5073, 2022.

\bibitem{chellapandi2023convergence}
V.~P. Chellapandi, A.~Upadhyay, A.~Hashemi, and S.~H. {\.Z}ak, ``{On the
  Convergence of Decentralized Federated Learning Under Imperfect Information
  Sharing},'' \emph{IEEE Control Systems Letters}, 2023.

\bibitem{chellapandi2023survey}
V.~P. Chellapandi, L.~Yuan, S.~H. Zak, and Z.~Wang, ``{A Survey of Federated
  Learning for Connected and Automated Vehicles},'' \emph{2023 IEEE 26th
  International Conference on Intelligent Transportation Systems (ITSC)}, 2023.

\bibitem{chellapandi2023federated}
V.~P. Chellapandi, L.~Yuan, C.~G. Brinton, S.~H. Zak, and Z.~Wang, ``{Federated
  Learning for Connected and Automated Vehicles: A Survey of Existing
  Approaches and Challenges},'' \emph{IEEE Transactions on Intelligent
  Vehicles}, 2023.

\bibitem{yuan2023peer}
L.~Yuan, Y.~Ma, L.~Su, and Z.~Wang, ``Peer-to-peer federated continual learning
  for naturalistic driving action recognition,'' in \emph{Proceedings of the
  IEEE/CVF Conference on Computer Vision and Pattern Recognition}, 2023, pp.
  5249--5258.

\bibitem{qi2023fl}
P.~Qi, D.~Chiaro, and F.~Piccialli, ``{FL-FD: Federated learning-based fall
  detection with multimodal data fusion},'' \emph{Information Fusion}, p.
  101890, 2023.

\bibitem{xiong2022unified}
B.~Xiong, X.~Yang, F.~Qi, and C.~Xu, ``A unified framework for multi-modal
  federated learning,'' \emph{Neurocomputing}, vol. 480, pp. 110--118, 2022.

\bibitem{feng2023fedmultimodal}
T.~Feng, D.~Bose, T.~Zhang, R.~Hebbar, A.~Ramakrishna, R.~Gupta, M.~Zhang,
  S.~Avestimehr, and S.~Narayanan, ``Fedmultimodal: A benchmark for multimodal
  federated learning,'' \emph{arXiv preprint arXiv:2306.09486}, 2023.

\bibitem{salehi2022flash}
B.~Salehi, J.~Gu, D.~Roy, and K.~Chowdhury, ``Flash: Federated learning for
  automated selection of high-band mmwave sectors,'' in \emph{IEEE INFOCOM
  2022-IEEE Conference on Computer Communications}.\hskip 1em plus 0.5em minus
  0.4em\relax IEEE, 2022, pp. 1719--1728.

\bibitem{chen2022fedmsplit}
J.~Chen and A.~Zhang, ``Fedmsplit: Correlation-adaptive federated multi-task
  learning across multimodal split networks,'' in \emph{Proceedings of the 28th
  ACM SIGKDD Conference on Knowledge Discovery and Data Mining}, 2022, pp.
  87--96.

\bibitem{imteaj2021survey}
A.~Imteaj, U.~Thakker, S.~Wang, J.~Li, and M.~H. Amini, ``A survey on federated
  learning for resource-constrained iot devices,'' \emph{IEEE Internet of
  Things Journal}, vol.~9, no.~1, pp. 1--24, 2021.

\bibitem{fang2023submodel}
W.~Fang, D.-J. Han, and C.~G. Brinton, ``Submodel partitioning in hierarchical
  federated learning: Algorithm design and convergence analysis,'' \emph{arXiv
  preprint arXiv:2310.17890}, 2023.

\bibitem{sundararajan2020many}
M.~Sundararajan and A.~Najmi, ``The many shapley values for model
  explanation,'' in \emph{International conference on machine learning}.\hskip
  1em plus 0.5em minus 0.4em\relax PMLR, 2020, pp. 9269--9278.

\bibitem{lundberg2017unified}
S.~M. Lundberg and S.-I. Lee, ``A unified approach to interpreting model
  predictions,'' \emph{Advances in neural information processing systems},
  vol.~30, 2017.

\bibitem{lundberg2020local2global}
S.~M. Lundberg, G.~Erion, H.~Chen, A.~DeGrave, J.~M. Prutkin, B.~Nair, R.~Katz,
  J.~Himmelfarb, N.~Bansal, and S.-I. Lee, ``From local explanations to global
  understanding with explainable ai for trees,'' \emph{Nature Machine
  Intelligence}, vol.~2, no.~1, pp. 2522--5839, 2020.

\bibitem{yuan2022interpretable}
L.~Yuan, J.~Andrews, H.~Mu, A.~Vakil, R.~Ewing, E.~Blasch, and J.~Li,
  ``Interpretable passive multi-modal sensor fusion for human identification
  and activity recognition,'' \emph{Sensors}, vol.~22, no.~15, p. 5787, 2022.

\bibitem{yuan2023three}
L.~Yuan, H.~Chen, R.~Ewing, E.~Blasch, and J.~Li, ``Three dimensional indoor
  positioning based on passive radio frequency signal strength distribution,''
  \emph{IEEE Internet of Things Journal}, vol.~10, no.~15, pp.
  13\,933--13\,944, 2023.

\bibitem{yuan2023passive}
L.~Yuan, H.~Chen, R.~Ewing, and J.~Li, ``Passive radio frequency-based 3d
  indoor positioning system via ensemble learning,'' \emph{arXiv preprint
  arXiv:2304.06513}, 2023.

\bibitem{yang2023extraction}
S.~Yang, L.~Yuan, and J.~Li, ``Extraction and denoising of human signature on
  radio frequency spectrums,'' in \emph{2023 IEEE International Conference on
  Consumer Electronics (ICCE)}.\hskip 1em plus 0.5em minus 0.4em\relax IEEE,
  2023, pp. 1--6.

\bibitem{shah2021model}
S.~M. Shah and V.~K. Lau, ``Model compression for communication efficient
  federated learning,'' \emph{IEEE Transactions on Neural Networks and Learning
  Systems}, 2021.

\bibitem{yuan2023federated}
L.~Yuan, L.~Su, and Z.~Wang, ``Federated transfer-ordered-personalized learning
  for driver monitoring application,'' \emph{IEEE Internet of Things Journal},
  vol.~10, no.~20, pp. 18\,292--18\,301, May 2023.

\bibitem{wu2022communication}
C.~Wu, F.~Wu, L.~Lyu, Y.~Huang, and X.~Xie, ``Communication-efficient federated
  learning via knowledge distillation,'' \emph{Nature communications}, vol.~13,
  no.~1, p. 2032, 2022.

\bibitem{delpreto2022actionsense}
J.~DelPreto, C.~Liu, Y.~Luo, M.~Foshey, Y.~Li, A.~Torralba, W.~Matusik, and
  D.~Rus, ``Actionsense: A multimodal dataset and recording framework for human
  activities using wearable sensors in a kitchen environment,'' \emph{Advances
  in Neural Information Processing Systems}, vol.~35, pp. 13\,800--13\,813,
  2022.

\end{thebibliography}

\end{document}